\documentclass[runningheads]{llncs}
\usepackage[T1]{fontenc}
\usepackage{graphicx}
\usepackage{tcolorbox}
\usepackage{subcaption} 
\usepackage{color}
\usepackage{amsmath}
\usepackage{subcaption} 
\begin{document}
\title{Rigidity in LLM Bandits with Implications for Human-AI Dyads}


\author{
Haomiaomiao Wang\inst{1}\orcidID{0009-0005-5961-1847} 
\and
Tomás E Ward\inst{1,2}\orcidID{0000-0002-6173-6607}
\and
Lili Zhang\inst{1,2}\orcidID{0000-0002-2203-2949}
}

\authorrunning{H. Wang et al.}
\institute{Insight Research Ireland Centre for Data Analytics, Ireland\and School of Computing, Dublin City University, Ireland }
%
\maketitle              
\begin{abstract}
We test whether LLMs show robust decision biases. Treating models as participants in two-arm bandits, we ran 200×100 trials per condition across four decoding configurations. Under symmetric rewards, models amplified positional order into stubborn one-arm policies. Under asymmetric rewards, they exploited rigidly yet underperformed an oracle and rarely re-checked. The observed patterns were consistent across manipulations of temperature and top-p, with top-k held at the provider default, indicating that the qualitative behaviours are robust to the two decoding knobs typically available to practitioners. Crucially, moving beyond descriptive metrics to computational modelling, a hierarchical Rescorla-Wagner-softmax fit revealed the underlying strategies: low learning rates and very high inverse temperatures, which together explain both noise-to-bias amplification and rigid exploitation. These results position minimal bandits as a tractable probe of LLM decision tendencies and motivate hypotheses about how such biases could shape human-AI interaction.
\end{abstract}

\section{Introduction}
Large language models (LLMs) are increasingly embedded in interactive settings, where their outputs guide human choices \cite{zhaoRecommenderSystemsEra2024}. Recent work shows that when humans interact with biased AI systems, their own judgements can become more biased over time, often without realising the extent of the AI’s influence \cite{glickmanHowHumanAI2024}. This raises a critical gap: benchmark evaluations capture accuracy, but rarely reveal the decision tendencies LLMs bring to interactive context, or how those tendencies might shape human-AI dyads \cite{berrettaDefiningHumanAITeaming2023,zhangAdversarialTestingLLMs2025}.

To address this gap, we borrow tools from cognitive science. Critics often argue that applying cognitive models to LLMs is misguided, given their weak architectural similarity to the brain and stochastic decoding \cite{benderDangersStochasticParrots2021}. Yet two considerations support the approach. First, theory-driven models often prove useful before mechanistic explanations are complete. Boltzmann’s “logical jumps” in statistical physics exemplify how bold, top-down reasoning can yield correct predictions despite incomplete foundations \cite{rosasTopdownBottomupNeuroscience2025}. Likewise, cognitive tasks may reveal meaningful functional alignments between LLM behaviour and human decision patterns, even without structural equivalence. Second, cognition is not only internal: Clark \cite{clarkExtendingMindsGenerative2025} argues that minds extend into tools, environments, and increasingly generative AI systems. From this relational view, LLMs participate in cognitive processes by shaping user judgments, choices, and beliefs. Thus, cognitive theory offers a principled lens for probing LLM decision tendencies,  even if they do not “possess” cognition in a biological sense.

In cognitive psychology, bandits provide a minimal and interpretable probe of bias and control \cite{schulzFindingStructureMultiarmed2020}. They dissociate action preference under ambiguity, learning and exploitation when one option is superior, and flexibility after feedback. Treating an LLM as a “participant” in this paradigm lets us measure constructs such as choice bias, stubbornness, exploration, and rigidity without heavy task semantics \cite{cheungLargeLanguageModels2025}. If these tendencies prove robust, they may be precisely the kinds of biases that spill over when LLMs act as advisors \cite{huangRecommenderAIAgent2025}.

\section{Methods}
\subsection{Experimental Design} 

We evaluated DeepSeek, GPT-4.1, and Gemini-2.5 (API versions listed in the repository \cite{llmrigidity2025}) across a large set of simulated bandit experiments. For each model we ran $N = 200$ independent simulations per condition, with $T = 100$ trials per run. The factorial design crossed symmetric and asymmetric reward structures with four decoding configurations. 

In the symmetric condition, both arms had equal reward probabilities ($p_X = 0.25,\; p_Y = 0.25$), with unbiased choices that should approximate a 50/50 split. In the asymmetric condition, one arm was superior ($p_X = 0.75,\; p_Y = 0.25$), requiring models to balance exploitation of the better option with flexibility to re-check the inferior one. The four conditions were governed by two decoding parameters. We manipulated \emph{temperature} and \emph{top-$p$} while leaving top-$k$ fixed at the provider default, defining four conditions (Table \ref{tab:decoding_regimes}). Temperature scales logits prior to sampling, with higher values producing more entropy \cite{renzeEffectSamplingTemperature2024}, while top-$p$ restricts sampling to the smallest probability mass $\geq p$ \cite{nguyenTurningHeatMinp2025}, with higher values including more of the probability tail.

\begin{table}[h!]
\setlength{\abovecaptionskip}{0pt}   
\setlength{\belowcaptionskip}{0pt}   
\setlength{\intextsep}{0pt}          
\centering
\caption{Decoding Configurations with Temperature and Top-$p$ Settings}
\label{tab:decoding_regimes}
\begin{tabular}{|l|l|l|}
\hline
\textbf{Strategy} & \textbf{Temperature} & \textbf{Top-$p$} \\
\hline
Strict        & 0.0 & 0.5 \\
Moderate      & 1.0 & 0.5 \\
Default-like  & 1.0 & 1.0 \\
Exploratory   & 2.0 & 1.0 \\
\hline
\end{tabular}
\end{table}

The experiment interaction with a fixed message structure is below. To enforce a categorical response, we set \texttt{max\_tokens = 1} and parsed a single character. This ensured that choices remained strictly binary. If the returned token was not exactly $X$ or $Y$, we treated the response as an invalid choice. Invalid choices were coded as the failure option with reward set to $0$ and included in analyses, and the overall invalid rate is reported separately. Condition-level invalid-response rates are included in the public repository.

\begin{tcolorbox}[title=Example Prompt, colback=gray!5!white, colframe=black!75!black]
\textbf{System:} You are a space explorer in a game. 
Your task is to choose between visiting Planet X or Planet Y in each round, 
aiming to find as many gold coins as possible. 
The probability of finding gold coins on each planet is unknown at the start, 
but you can learn and adjust your strategy based on the outcomes of your previous visits. 
Respond with `X' for Planet X or `Y' for Planet Y.

\medskip
\textbf{Prompt:} Your previous space travels went as follows: \\
- In Trial 1, you went to Planet X and found 100 gold coins. \\
- In Trial 2, you went to Planet X and found nothing. \\
- In Trial 3, you went to Planet Y and found nothing. \\

Q: Which planet do you want to go to in Trial 4? \\
A: Planet
\end{tcolorbox}

Stimulus generation and logging were implemented in Python using provider chat APIs, analyses and visualization were carried out in R, and hierarchical inference was conducted in Stan. All per-run CSVs, condition-level summaries, analysis scripts, and Stan code are provided in a public repository \cite{llmrigidity2025}.

\subsection{Behavioural Indices and Statistical Summaries}

Analyses were conducted at the run level and then aggregated across runs per cell. We computed the metrics in Table \ref{tab:metrics}. 

For each cell we report run-level means with \(\pm\) 95\% confidence intervals across the 200 runs to quantify uncertainty across the 200 simulated participants. Because the behavioural indices are bounded and the data are hierarchically structured, classical significance tests are not well-suited. We therefore interpret differences between decoding strategies based on the magnitude of the observed effects and on whether their 95\% confidence intervals overlap, rather than through classical significance tests.
 
\begin{table}[h!]
\setlength{\abovecaptionskip}{0pt}   
\setlength{\belowcaptionskip}{0pt}   
\setlength{\intextsep}{0pt}          
\centering
\caption{Summary of Computed Behavioural Metrics}
\label{tab:metrics}
\begin{tabular}{|l|l|p{3.5cm} p{4cm}}
\hline
\textbf{Metric} & \textbf{Definition} \\
\hline
Total Reward & Sum over trials \cite{zhangAdversarialTestingLLMs2025}\\[4pt]
Target-arm Rate & Fraction choosing the higher-probability arm \cite{zhangAdversarialTestingLLMs2025} \\[4pt]
Loss-Shift Win-Shift & Probability of switching after a loss / win \cite{zhangAdversarialTestingLLMs2025} \\[4pt]
Choice Bias Index & $\bar{c} - 0.5,\;\; \bar{c} = P(\text{choose } Y)$ \cite{lopez-persemHowPriorPreferences2016}\\[6pt]
Stubbornness Rate & Fraction of runs with $\bar{c} \geq 0.8$ or $\bar{c} \leq 0.2$ \cite{hunterOptimizingOpinionsStubborn2022} \\[6pt]
Amplification Index & Fraction of post-warm-up runs that are monomorphic \cite{hoelzemannBanditsLab2021} \\[6pt]
Rigidity Index & $\ 1 - \overline{Loss-Shift}$ (post-warm-up) \cite{knepSocialAloofnessAssociated2025}\\[6pt]
Adjusted Choice Bias & \text{Target Rate} - 0.90 (only under the asymmetric reward structure)\\[4pt]
\hline
\end{tabular}
\end{table}

\subsection{Computational Modelling}
To explain the observed patterns mechanistically, we fit a hierarchical Rescorla--Wagner learning model with a softmax policy in Stan to each reward structure separately \cite{bariReinforcementLearningModeling2022}. For run \(i\), the chosen arm’s value updated as

\[
    V_{t+1}(a) = V_t(a) + A_i \big( r_t - V_t(a) \big),
\]
with learning rate \(A_i \in (0,1)\); the unchosen value was unchanged. 
Choice probability followed

\[
    P(Y_t = 1) = \text{logit}^{-1} \!\Big( \tau_i \, [\, V_t(Y) - V_t(X) \,] \Big),
\]
with inverse temperature \(\tau_i > 0\). Values were initialized at zero. Individual parameters \((A_i, \tau_i)\) were drawn via probit transforms from group-level normals 
with hyper-means \(\mu\) and scales \(\sigma\):  

\[
    \mu \sim \mathcal{N}(0,1), 
    \quad \sigma \sim \mathcal{N}^+(0,0.2).
\]

To ensure interpretability, \(\tau\) was bounded to \([0,5]\) on the natural scale 
by multiplying the probit output by 5. For group-level parameters \((\mu_A, \mu_\tau)\), we report posterior means and 95\% credible intervals, which provide the Bayesian measure of uncertainty in the inferred learning-rate and inverse-temperature estimates. The model produced group-level summaries \((\mu_A, \mu_\tau)\), per-run log-likelihoods, 
and posterior-predictive choices for model checking \cite{horvathHumanBeliefStatebased2021,sugawaraDissociationAsymmetricValue2021}.  

\section{Results}

\subsection{Behavioural Metrics}

\begin{figure}[htbp]
\setlength{\abovecaptionskip}{0pt}
\setlength{\belowcaptionskip}{0pt}
\setlength{\intextsep}{0pt}
\centering
\includegraphics[width=1\linewidth]{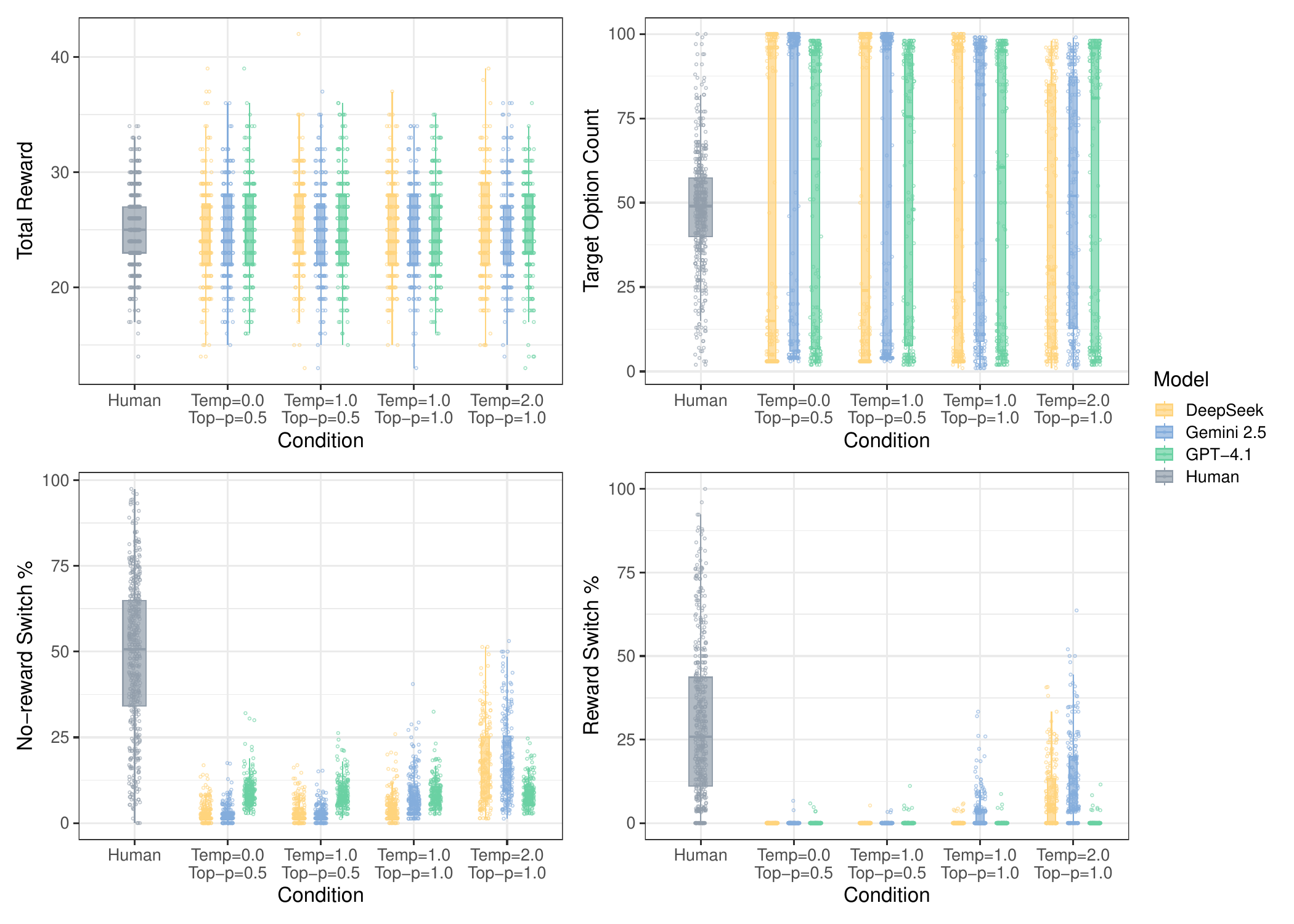}
\caption{Symmetric Bandit Behavioural Metrics}
\label{fig:symetric}
\end{figure}

With equal reward probability, an unbiased learner should split choices evenly, 
yielding $\approx 25$ rewards per 100 trials and target rates near $0.50$. 
Humans matched this benchmark: (X,Y) = ($0.49 \pm 0.18$, $0.51 \pm 0.18$), with chance‑level totals \cite{danChoiceArchitectureChoice2019} (Fig.~\ref{fig:symetric}). LLMs, however, departed systematically. Total rewards were near chance 
(e.g., DeepSeek Temp=0.0, Top‑p=0.5: $24.60 \pm 0.62$; 
Gemini‑2.5 Temp=0.0, Top‑p=0.5: $24.71 \pm 0.56$; 
GPT‑4.1 Temp=0.0, Top‑p=0.5: $25.38 \pm 0.61$), but choice distributions diverged. 
On the first trial models almost always chose X, and because both arms occasionally pay, 
early X wins reinforced persistence. Gemini‑2.5 showed the strongest X‑tilt in the stric decoding strategy
(Temp=0.0, Top‑p=0.5: (X,Y)=($0.61 \pm 0.44$, $0.39 \pm 0.44$)), 
which attenuated under exploratory strategy (Temp=2.0, Top‑p=1.0: (X,Y)=($0.50 \pm 0.36$, $0.50 \pm 0.36$)). 
DeepSeek stayed closer to even (e.g., Temp=1.0, Top‑p=0.5: (X,Y)=($0.50 \pm 0.45$, $0.50 \pm 0.45$)) 
but could flip toward Y at high temperature (Temp=2.0, Top‑p=1.0: (X,Y)=($0.44 \pm 0.37$, $0.56 \pm 0.37$)). 
GPT‑4.1 leaned toward X (e.g., Temp=1.0, Top‑p=0.5: (X,Y)=($0.55 \pm 0.40$, $0.45 \pm 0.40$)), though less than Gemini.

Switching behaviour reinforced this picture: Loss‑Shift was near zero in the strict strategy 
(DeepSeek $0.03 \pm 0.00$; Gemini‑2.5 $0.03 \pm 0.00$; GPT‑4.1 $0.09 \pm 0.01$), 
and Win‑Shift was essentially absent, rising only with exploration 
(DeepSeek Temp=2.0, Top‑p=1.0: $0.09 \pm 0.01$; Gemini‑2.5 Temp=2.0, Top‑p=1.0: $0.13 \pm 0.02$). Bias indices converged on the same message. Mean Choice Bias Index deviated from $0$ 
(Gemini‑2.5 most negative: $-0.11 \pm 0.06$; GPT‑4.1: $-0.03$ to $-0.05 \pm 0.06$; DeepSeek: $0.03 \pm 0.06$). 
Stubbornness Rate was high in the strict strategy (DeepSeek $0.97 \pm 0.02$, 
Gemini‑2.5 $0.95 \pm 0.03$, GPT‑4.1 $0.90 \pm 0.04$ at Temp=0.0, Top‑p=0.5). 
Amplification Index was large for DeepSeek and Gemini‑2.5 (0.62-0.67 $\pm 0.07$) and lower for GPT‑4.1 (0.33-0.43 $\pm 0.07$). 
Rigidity Index hovered near ceiling (0.96-0.99 $\pm 0.01$) except under exploratory decoding 
(e.g., DeepSeek Temp=2.0, Top‑p=1.0: $0.85 \pm 0.02$). In ambiguity, models amplify a positional nudge into stubborn choice.

\begin{figure}[htbp]
\setlength{\abovecaptionskip}{0pt}
\setlength{\belowcaptionskip}{0pt}
\setlength{\intextsep}{0pt}
\centering
\includegraphics[width=1\linewidth]{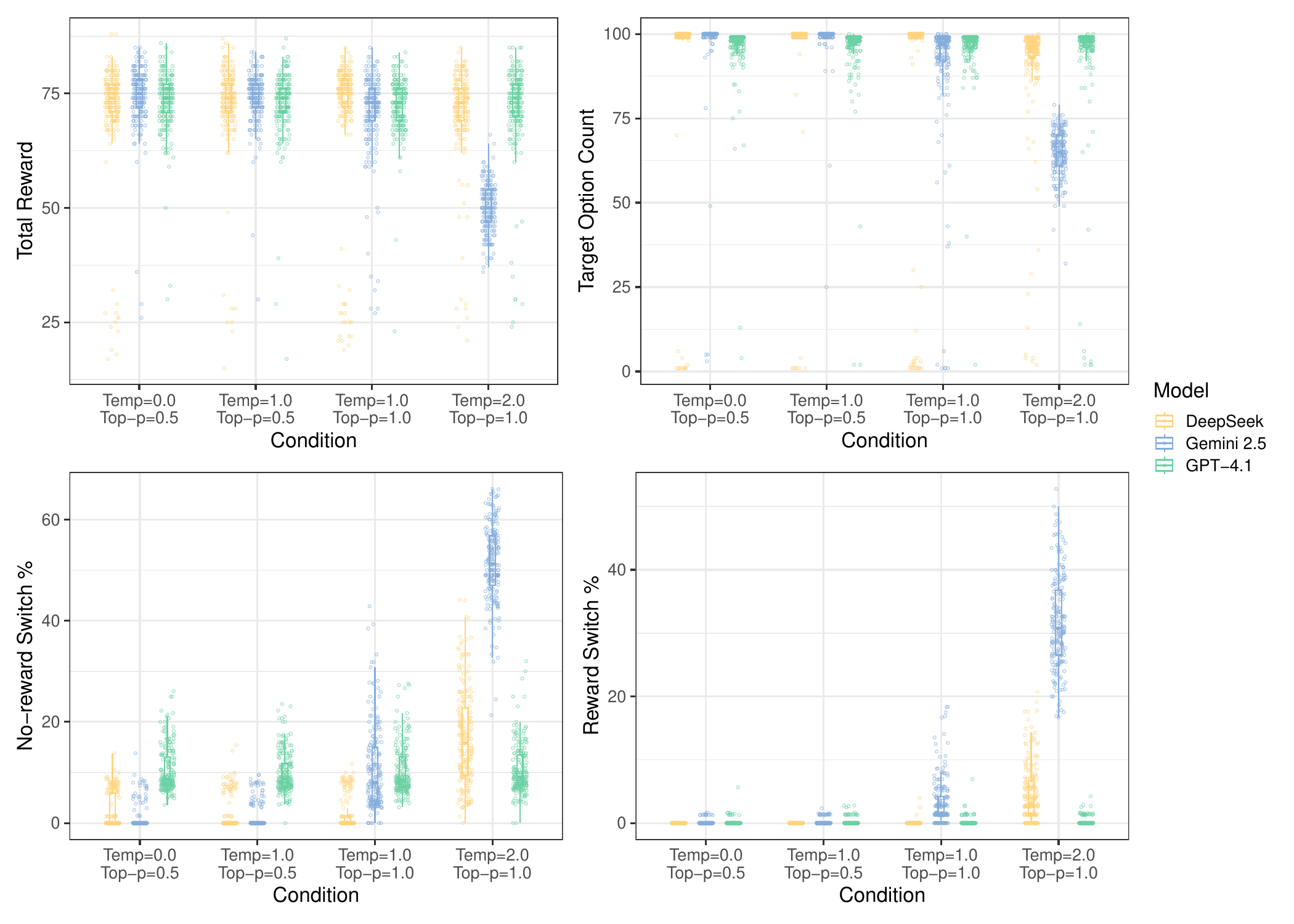}
\caption{Asymmetric Bandit Behavioural Metrics}
\label{fig:asymetric}
\end{figure}

With $p_X=0.75,\;p_Y=0.25$, an optimal learner should approach 75 rewards and a target rate near 1.0. 
LLMs converged to the better arm but did so rigidly. First‑trial X bias again appeared and, here, was reward‑consistent, 
accelerating early convergence. Totals clustered below the oracle yet high for DeepSeek and GPT‑4.1 
(DeepSeek Temp=1.0, Top‑p=0.5: $72.68 \pm 1.44$; GPT‑4.1 Temp=0.0, Top‑p=0.5: $73.15 \pm 0.92$) (Fig.~\ref{fig:asymetric}). Gemini‑2.5 peaked under strict decoding ($74.22 \pm 1.00$ at Temp=0.0, Top‑p=0.5) 
but collapsed under exploration ($50.06 \pm 0.73$ at Temp=2.0, Top‑p=1.0), due to around $2\%$ invalid outputs. 

Target‑arm rates were near‑ceiling in the strict strategy 
(DeepSeek $0.95 \pm 0.03$, Gemini‑2.5 $0.98 \pm 0.02$, GPT‑4.1 $0.96 \pm 0.01$) 
and dropped with exploration (Gemini Temp=2.0, Top‑p=1.0: $0.65 \pm 0.01$). 
Adjusted Choice Bias Index values were negative ($-0.04$ to $-0.09$), indicating under‑selection relative to an oracle.

Switching patterns confirmed rigidity: Loss-Shift stayed minimal in the strict strategy 
(DeepSeek $0.02 \pm 0.01$, Gemini‑2.5 $0.01 \pm 0.00$), moderate for GPT‑4.1 ($0.10 \pm 0.01$), 
and spiked for exploratory Gemini (Temp=2.0, Top‑p=1.0: $0.51 \pm 0.01$). 
Win-Shift was near zero except under exploration (DeepSeek $0.05 \pm 0.01$; Gemini $0.32 \pm 0.01$). 
Stubbornness and Amplification exceeded $0.90$ in most settings but collapsed for Gemini at Temp=2.0, Top‑p=1.0. 
Rigidity Index was near ceiling for DeepSeek (0.999-1.000 $\pm 0.001$) and GPT‑4.1 (0.93-0.94 $\pm 0.01$), 
dropping to $0.48 \pm 0.01$ for exploratory Gemini. 
With a clear winner, models exploit hard and re‑check little; heavy exploration degrades efficiency.

\subsection{Computational Modelling}

We fit a hierarchical Rescorla--Wagner model with a softmax policy in Stan to explain why LLMs turn ambiguity into
stubborn choice and clarity into rigid exploitation. The hierarchy places per-run parameters $(A_i,\tau_i)$ under
group‑level hyperparameters $(\mu_A,\mu_\tau,\sigma_A,\sigma_\tau)$, so that $\mu_A$ and $\mu_\tau$ summarise the typical learning rate and inverse temperature. Uncertainty in the inferred parameters is expressed through posterior means and 95\% credible intervals, which quantify the range of learning-rate and inverse-temperature values compatible with the data.The learning rate $A\in[0,1]$ controls how strongly prediction errors update values: higher $A$ means faster
adaptation. The inverse temperature $\tau\ge 0$ controls choice determinism: higher $\tau$ means more
deterministic softmax, then $\tau\!\to\!\infty$ approximates greedy choice.

\begin{figure}[h!]
\centering
\begin{subfigure}{\linewidth}
  \centering
  \includegraphics[width=\linewidth]{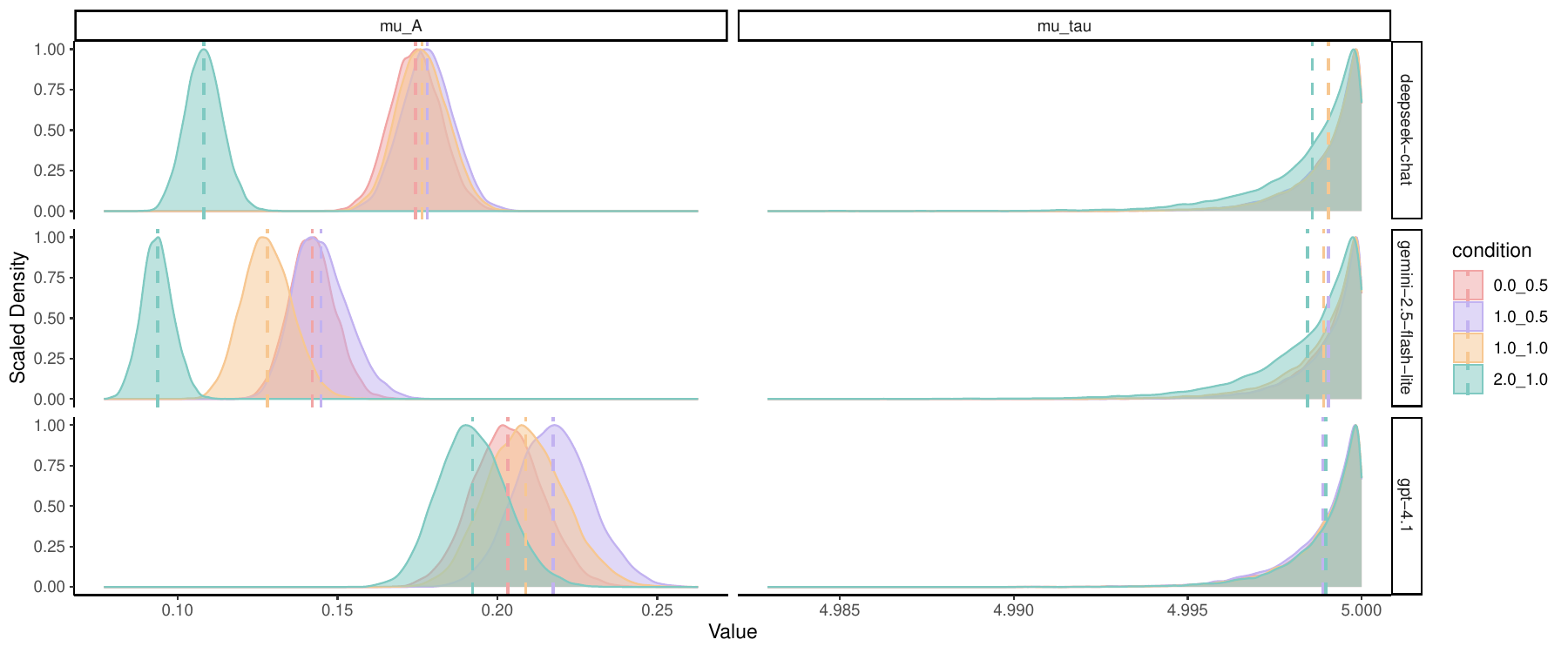}
  \caption{Symmetric bandit: group‑level posteriors for $\mu_A$ and $\mu_\tau$.}
  \label{fig:fig3a}
\end{subfigure}
\vspace{0.5em}
\begin{subfigure}{\linewidth}
  \centering
  \includegraphics[width=\linewidth]{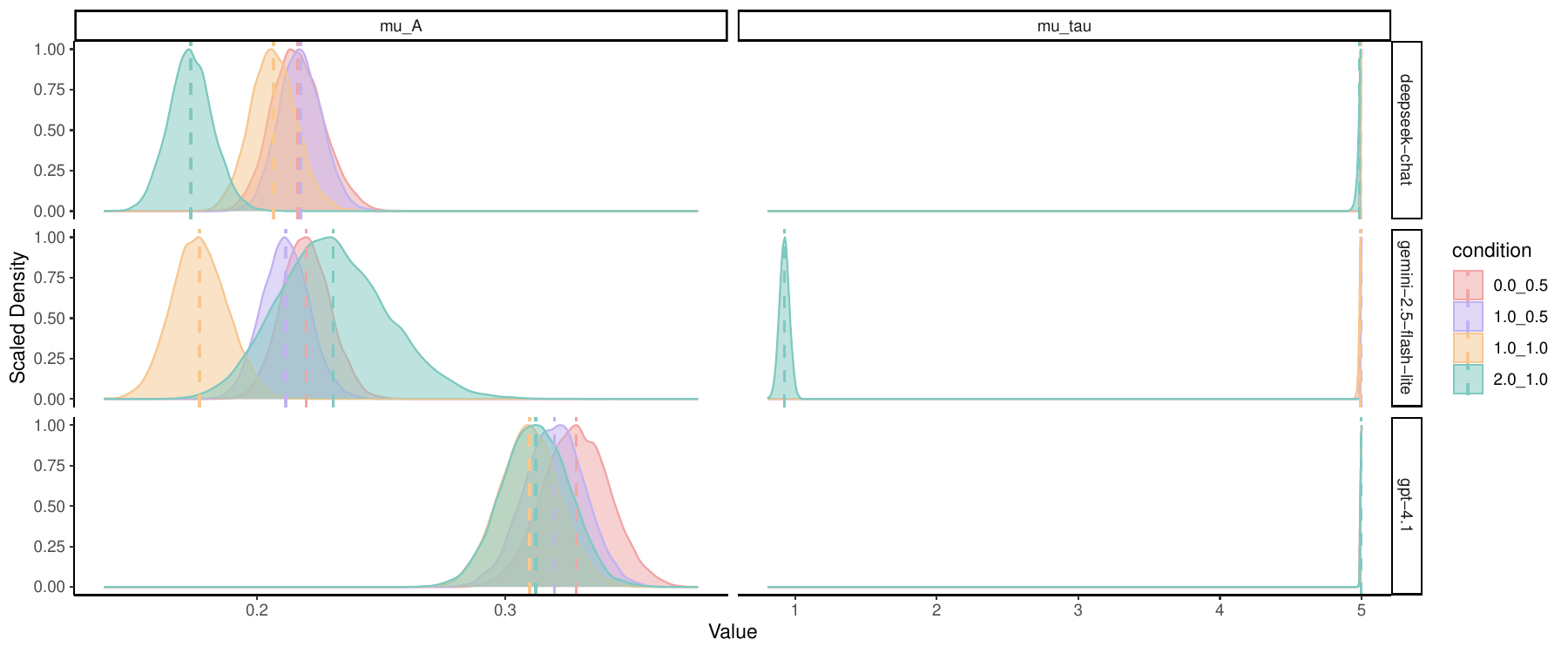}
  \caption{Asymmetric bandit: group‑level posteriors for $\mu_A$ and $\mu_\tau$}
  \label{fig:fig3b}
\end{subfigure}
\caption{Posterior Densities of Group‑level Parameters}
\label{fig:fig3}
\end{figure}

Across the symmetric settings, the group learning rates were uniformly low $\mu_A \in [0.09,\,0.22]$ (Fig. \ref{fig:fig3a}). The group inverse temperatures were effectively at the ceiling $\mu_\tau \in [4.9984,\,4.9991]$. This reflects genuine over-determinism in policy rather than a boundary-induced distortion. Slow updating paired with near‑deterministic choice causes early fluctuations to be entrenched, matching the high stubbornness/low switching observed under symmetry. For the asymmetric counterparts, learning rates increased $\mu_A \in [0.17,\,0.33]$ (Fig. \ref{fig:fig3b}). Inverse temperatures again clustered near the ceiling $\mu_\tau \in [4.991,\,4.998]$,
with two notable deviations: DeepSeek at $T\!=\!2.0$, $p\!=\!1.0$ dipped slightly ($\mu_\tau\!\approx\!4.986$), and
Gemini at $T\!=\!2.0$, $p\!=\!1.0$ collapsed ($\mu_\tau\!\approx\!0.93$), consistent with that cell’s high switching and
invalid outputs. Overall, the same strategy, low $\mu_A$ with very high $\mu_\tau$, accounts for near‑deterministic exploitation of the better arm and the reluctance to re‑check.

\begin{figure}[h!]
\centering
\includegraphics[width=\linewidth]{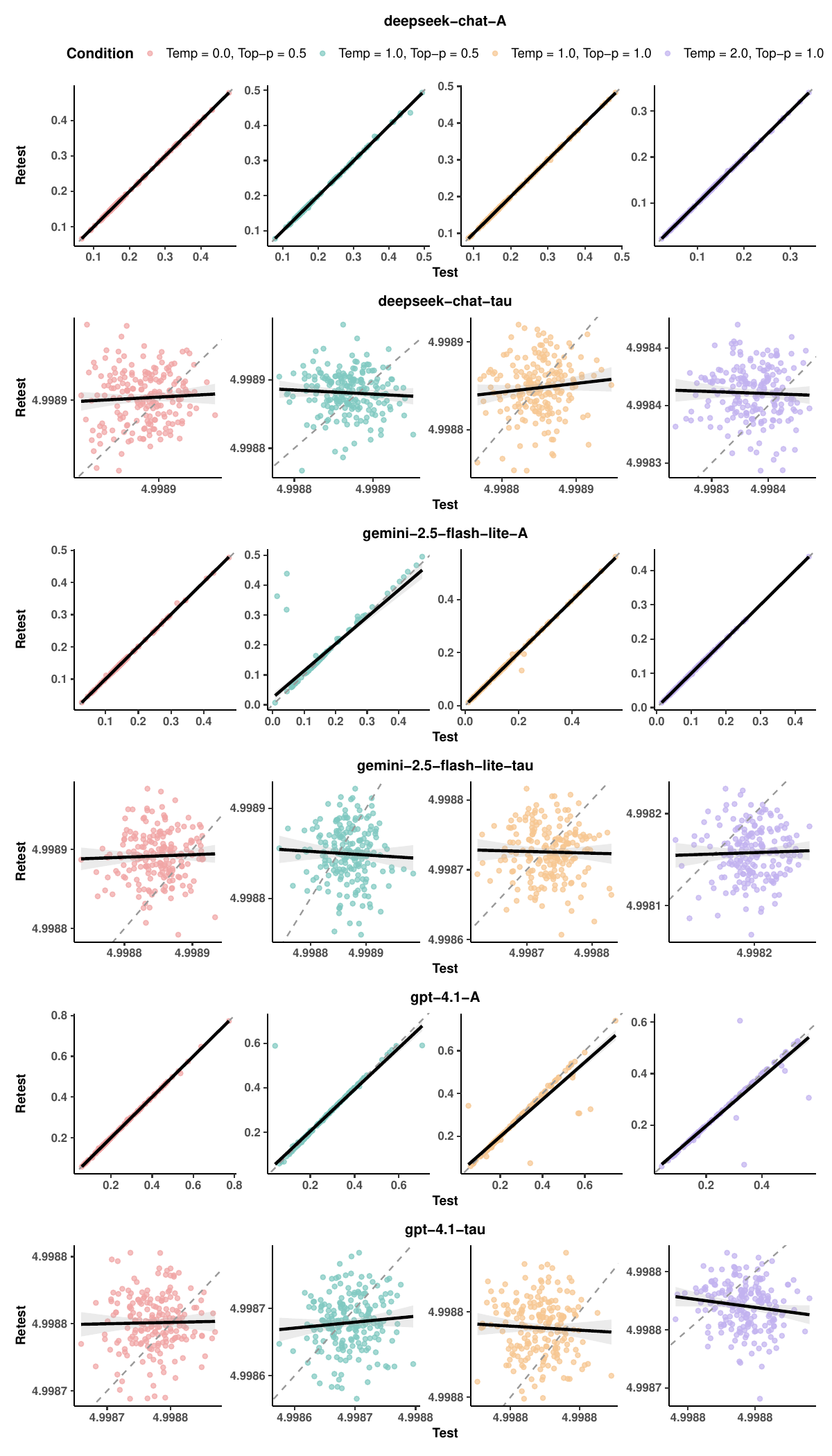}
\caption{Test-retest Reliability on the Symmetric Bandit}
\label{fig:icc_patchwork}
\end{figure}

\begin{figure}[h!]
\centering
\includegraphics[width=\linewidth]{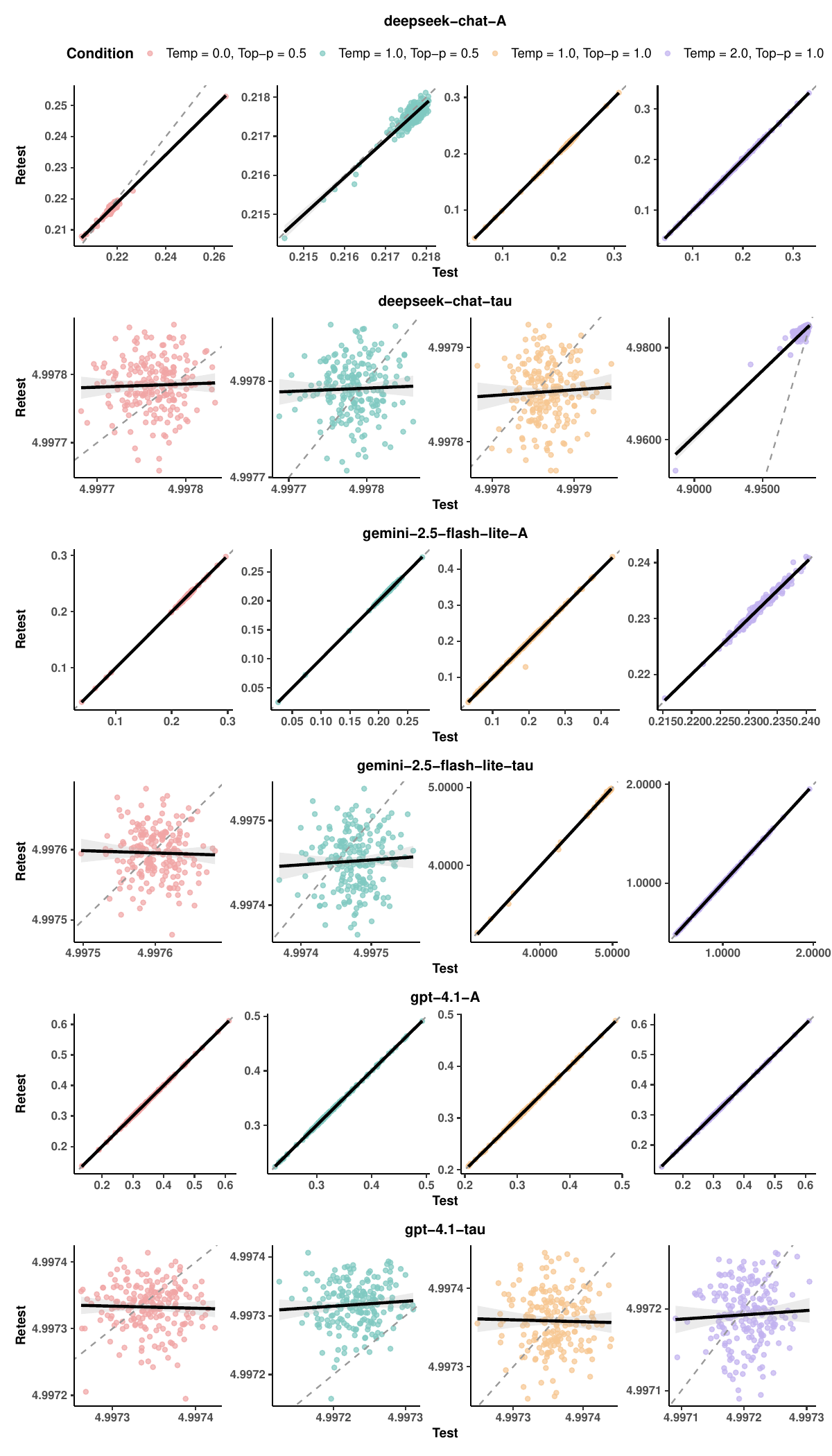}
\caption{Test-retest Reliability on the Asymmetric Bandit}
\label{fig:icc_patchwork_1}
\end{figure}

We computed ICC(3,1) on per-run posterior means \((A_i,\tau_i)\) for run-level reliability. Learning rate \(A\) was highly reliable across models and decoders, whereas inverse temperature showed \(\mathrm{ICC}(\tau)\!\approx\!0\), consistent with range restriction from ceiling saturation rather than noisy estimation (Fig.~\ref{fig:icc_patchwork}). With a reward gradient, \(A\) remained very reliable for all models, while reliability for \(\tau\) bifurcated by decoder and model (Fig.~\ref{fig:icc_patchwork_1}). 

\section{Facts about LLM Decision Policies}

Our two tasks make the exploration-exploitation dilemma explicit. The symmetric bandit ($0.25/0.25$) is a low‑opportunity environment: rewards are sparse and the value of information is high, so flexible exploration is essential. The asymmetric bandit ($0.75/0.25$) is a high‑opportunity environment: exploitation pays, but occasional re‑checks hedge against false certainty. However, LLMs allocate their ``exploration budget'' poorly: too little when information is valuable, and still too little when exploitation is warranted but periodic verification would improve efficiency.

Moreover, LLM agents are parameter-locked: they learn slowly and choose deterministically. Thus, raising temperature or top-\(p\) mainly changes the appearance of behaviour, while the underlying low-\(A\), high-\(\tau\) strategy persists. This underscores a practical point: adding sampling noise may surface format‑compliance failures rather than genuine epistemic exploration.

\section{Implications for theory and for human-AI dyads}

Our findings expose a form of internal bias amplification within LLM adaptive patterns: small, incidental asymmetries are not dampened by stochasticity but reinforced into stable, policy-level preferences. More fundamentally, LLMs appear opportunity-blind, which conserves exploration where uncertainty makes information most valuable and over-committing when clarity renders exploration cheap. This asymmetry constitutes a resource-allocation failure: exploration is not tuned to expected information gain. Instead of dynamically adjusting effort to environmental opportunity, LLMs default to a single, efficiency-oriented strategy that treats uncertainty as noise to be eliminated rather than as information to be harvested. The result is a form of epistemic inertia, where early preferences tend to persist because new evidence has little influence.

For human-AI dyads, these adaptive biases may carry direct risks. Deterministic, confident advice can amplify early cues into unwarranted certainty, leading to false positives under ambiguity, premature commitment to an unverified option, and false negatives under clarity, failure to revisit rare but consequential alternatives. Order effects in prompts act as a form of choice architecture that shapes model output, and could influence user reasoning when models are used as advisors. Higher-temperature decoding increases behavioural variability but also raises the rate of format errors, making it harder to distinguish exploration from simple output instability. In applied contexts, such tendencies could translate into positional bias or premature lock-in when users rely on model advice. Such dyads may appear efficient but can be vulnerable if users mistake deterministic output for correctness \cite{challenArtificialIntelligenceBias2019,nord-bronzykAssessingRiskImplementing2025}.

\section{Future directions}
The two‑arm bandit exposes clear regularities with minimal confounds, but richer tasks are needed to test boundary conditions and mitigations. On the task side, contextual and non‑stationary bandits can raise the value of information dynamically, probing whether models can direct exploration when stability is not rewarded. Social decision paradigms such as multi‑round trust tasks can test adaptation to strategic feedback. Prompt architecture should be systematically varied to quantify positional effects.

On the modelling side, extending beyond bounded‑$\tau$ softmax to include perseveration, lapse/format‑error channels, or uncertainty‑aware policies may reveal whether the observed ceiling on $\tau$ reflects genuine over‑determinism or model misspecification. Finally, moving from internal behaviour to communicative impact, experiments should measure how advice wording mediates bias transfer to humans, e.g., comparing biased vs.\ randomized advisors in controlled dyads.


%
%
%
\bibliographystyle{splncs04}
\bibliography{bib}

@article{bariReinforcementLearningModeling2022,
  title = {Reinforcement Learning Modeling Reveals a Reward-History-Dependent Strategy Underlying Reversal Learning in Squirrel Monkeys},
  author = {Bari, Bilal A. and Moerke, Megan J. and Jedema, Hank P. and Effinger, Devin P. and Cohen, Jeremiah Y. and Bradberry, Charles W.},
  year = {2022},
  month = feb,
  journal = {Behavioral Neuroscience},
  volume = {136},
  number = {1},
  pages = {46--60},
  issn = {1939-0084},
  doi = {10.1037/bne0000492},
  langid = {english},
  pmcid = {PMC8863624},
  pmid = {34570556}
}

@inproceedings{benderDangersStochasticParrots2021,
  title = {On the Dangers of Stochastic Parrots: Can Language Models Be Too Big?},
  shorttitle = {On the Dangers of Stochastic Parrots},
  booktitle = {Proceedings of the 2021 {{ACM Conference}} on {{Fairness}}, {{Accountability}}, and {{Transparency}}},
  author = {Bender, Emily M. and Gebru, Timnit and {McMillan-Major}, Angelina and Shmitchell, Shmargaret},
  year = {2021},
  month = mar,
  pages = {610--623},
  publisher = {ACM},
  address = {Virtual Event Canada},
  doi = {10.1145/3442188.3445922},
  urldate = {2025-10-02},
  isbn = {978-1-4503-8309-7},
  langid = {english}
}

@article{berrettaDefiningHumanAITeaming2023,
  title = {Defining Human-{{AI}} Teaming the Human-Centered Way: A Scoping Review and Network Analysis},
  shorttitle = {Defining Human-{{AI}} Teaming the Human-Centered Way},
  author = {Berretta, Sophie and Tausch, Alina and Ontrup, Greta and Gilles, Bj{\"o}rn and Peifer, Corinna and Kluge, Annette},
  year = {2023},
  month = sep,
  journal = {Frontiers in Artificial Intelligence},
  volume = {6},
  pages = {1250725},
  issn = {2624-8212},
  doi = {10.3389/frai.2023.1250725},
  urldate = {2025-10-02},
  langid = {english}
}

@article{cheungLargeLanguageModels2025,
  title = {Large Language Models Show Amplified Cognitive Biases in Moral Decision-Making},
  author = {Cheung, Vanessa and Maier, Maximilian and Lieder, Falk},
  year = {2025},
  month = jun,
  journal = {Proceedings of the National Academy of Sciences},
  volume = {122},
  number = {25},
  pages = {e2412015122},
  issn = {0027-8424, 1091-6490},
  doi = {10.1073/pnas.2412015122},
  urldate = {2025-10-02},
  langid = {english}
}

@article{clarkExtendingMindsGenerative2025,
  title = {Extending Minds with Generative {{AI}}},
  author = {Clark, Andy},
  year = {2025},
  month = may,
  journal = {Nature Communications},
  volume = {16},
  number = {1},
  pages = {4627},
  issn = {2041-1723},
  doi = {10.1038/s41467-025-59906-9},
  urldate = {2025-10-02},
  langid = {english}
}

@article{danChoiceArchitectureChoice2019,
  title = {From Choice Architecture to Choice Engineering},
  author = {Dan, Ohad and Loewenstein, Yonatan},
  year = {2019},
  month = jun,
  journal = {Nature Communications},
  volume = {10},
  number = {1},
  pages = {2808},
  issn = {2041-1723},
  doi = {10.1038/s41467-019-10825-6},
  urldate = {2025-10-03},
  langid = {english}
}

@article{glickmanHowHumanAI2024,
  title = {How Human--{{AI}} Feedback Loops Alter Human Perceptual, Emotional and Social Judgements},
  author = {Glickman, Moshe and Sharot, Tali},
  year = {2024},
  month = dec,
  journal = {Nature Human Behaviour},
  volume = {9},
  number = {2},
  pages = {345--359},
  issn = {2397-3374},
  doi = {10.1038/s41562-024-02077-2},
  urldate = {2025-10-02},
  langid = {english}
}

@article{hoelzemannBanditsLab2021,
  title = {Bandits in the Lab},
  author = {Hoelzemann, Johannes and Klein, Nicolas},
  year = {2021},
  journal = {Quantitative Economics},
  volume = {12},
  number = {3},
  pages = {1021--1051},
  issn = {1759-7323},
  doi = {10.3982/QE1389},
  urldate = {2025-10-03},
  copyright = {http://creativecommons.org/licenses/by-nc/4.0/},
  langid = {english}
}

@article{horvathHumanBeliefStatebased2021,
  title = {Human Belief State-Based Exploration and Exploitation in an Information-Selective Symmetric Reversal Bandit Task},
  author = {Horvath, Lilla and Colcombe, Stanley and Milham, Michael and Ray, Shruti and Schwartenbeck, Philipp and Ostwald, Dirk},
  year = {2021},
  month = dec,
  journal = {Computational Brain \& Behavior},
  volume = {4},
  number = {4},
  pages = {442--462},
  issn = {2522-0861, 2522-087X},
  doi = {10.1007/s42113-021-00112-3},
  urldate = {2025-10-02},
  langid = {english}
}

@article{huangRecommenderAIAgent2025,
  title = {Recommender {{AI}} Agent: Integrating Large Language Models for Interactive Recommendations},
  shorttitle = {Recommender {{AI}} Agent},
  author = {Huang, Xu and Lian, Jianxun and Lei, Yuxuan and Yao, Jing and Lian, Defu and Xie, Xing},
  year = {2025},
  month = jul,
  journal = {ACM Transactions on Information Systems},
  volume = {43},
  number = {4},
  pages = {1--33},
  issn = {1046-8188, 1558-2868},
  doi = {10.1145/3731446},
  urldate = {2025-10-02},
  langid = {english}
}

@misc{hunterOptimizingOpinionsStubborn2022,
  title = {Optimizing Opinions with Stubborn Agents},
  author = {Hunter, D. Scott and Zaman, Tauhid},
  year = {2022},
  month = jul,
  number = {arXiv:1806.11253},
  eprint = {1806.11253},
  primaryclass = {cs},
  publisher = {arXiv},
  doi = {10.48550/arXiv.1806.11253},
  urldate = {2025-10-03},
  archiveprefix = {arXiv},
  langid = {english}
}

@article{knepSocialAloofnessAssociated2025,
  title = {Social Aloofness Is Associated with Non-Social Explore-Exploit Decisions},
  author = {Knep, Evan and Yan, Xinyuan and Chen, Cathy S. and Jacob, Suma and Darrow, David P. and Ebitz, R. Becket and Grissom, Nicola and Herman, Alexander B.},
  year = {2025},
  month = jul,
  journal = {Communications Psychology},
  volume = {3},
  number = {1},
  pages = {106},
  issn = {2731-9121},
  doi = {10.1038/s44271-025-00278-7},
  langid = {english},
  pmcid = {PMC12263421},
  pmid = {40665112}
}

@article{lopez-persemHowPriorPreferences2016,
  title = {How Prior Preferences Determine Decision-Making Frames and Biases in the Human Brain},
  author = {{Lopez-Persem}, Aliz{\'e}e and Domenech, Philippe and Pessiglione, Mathias},
  year = {2016},
  month = nov,
  journal = {Elife},
  volume = {5},
  pages = {e20317},
  issn = {2050-084X},
  doi = {10.7554/eLife.20317},
  langid = {english},
  pmcid = {PMC5132340},
  pmid = {27864918}
}

@misc{nguyenTurningHeatMinp2025,
  title = {Turning up the Heat: Min-p Sampling for Creative and Coherent {{LLM}} Outputs},
  shorttitle = {Turning up the Heat},
  author = {Nguyen, Minh Nhat and Baker, Andrew and Neo, Clement and Roush, Allen and Kirsch, Andreas and {Shwartz-Ziv}, Ravid},
  year = {2025},
  month = jun,
  number = {arXiv:2407.01082},
  eprint = {2407.01082},
  primaryclass = {cs},
  publisher = {arXiv},
  doi = {10.48550/arXiv.2407.01082},
  urldate = {2025-10-03},
  archiveprefix = {arXiv},
  langid = {english}
}

@inproceedings{renzeEffectSamplingTemperature2024,
  title = {The Effect of Sampling Temperature on Problem Solving in Large Language Models},
  booktitle = {Findings of the {{Association}} for {{Computational Linguistics}}: {{EMNLP}} 2024},
  author = {Renze, Matthew},
  year = {2024},
  pages = {7346--7356},
  publisher = {Association for Computational Linguistics},
  address = {Miami, Florida, USA},
  doi = {10.18653/v1/2024.findings-emnlp.432},
  urldate = {2025-10-03},
  langid = {english}
}

@article{rosasTopdownBottomupNeuroscience2025,
  title = {Top-down and Bottom-up Neuroscience: Overcoming the Clash of Research Cultures},
  shorttitle = {Top-down and Bottom-up Neuroscience},
  author = {Rosas, Fernando E. and Luppi, Andrea I. and Mediano, Pedro A. M. and Kringelbach, Morten L. and Pessoa, Luiz and Turkheimer, Federico},
  year = {2025},
  month = sep,
  journal = {Nature Reviews Neuroscience},
  volume = {26},
  number = {9},
  pages = {513--515},
  issn = {1471-003X, 1471-0048},
  doi = {10.1038/s41583-025-00946-x},
  urldate = {2025-10-02},
  langid = {english}
}

@article{schulzFindingStructureMultiarmed2020,
  title = {Finding Structure in Multi-Armed Bandits},
  author = {Schulz, Eric and Franklin, Nicholas T. and Gershman, Samuel J.},
  year = {2020},
  month = jun,
  journal = {Cognitive Psychology},
  volume = {119},
  pages = {101261},
  issn = {00100285},
  doi = {10.1016/j.cogpsych.2019.101261},
  urldate = {2025-10-02},
  langid = {english}
}

@article{sugawaraDissociationAsymmetricValue2021,
  title = {Dissociation between Asymmetric Value Updating and Perseverance in Human Reinforcement Learning},
  author = {Sugawara, Michiyo and Katahira, Kentaro},
  year = {2021},
  month = feb,
  journal = {Scientific Reports},
  volume = {11},
  number = {1},
  pages = {3574},
  issn = {2045-2322},
  doi = {10.1038/s41598-020-80593-7},
  urldate = {2025-10-02},
  langid = {english}
}

@misc{zhangAdversarialTestingLLMs2025,
  title = {Adversarial Testing in {{LLMs}}: Insights into Decision-Making Vulnerabilities},
  shorttitle = {Adversarial Testing in {{LLMs}}},
  author = {Zhang, Lili and Wang, Haomiaomiao and Cheng, Long and Deng, Libao and Ward, Tomas},
  year = {2025},
  month = may,
  number = {arXiv:2505.13195},
  eprint = {2505.13195},
  primaryclass = {cs},
  publisher = {arXiv},
  doi = {10.48550/arXiv.2505.13195},
  urldate = {2025-10-02},
  archiveprefix = {arXiv},
  langid = {english}
}

@article{zhaoRecommenderSystemsEra2024,
  title = {Recommender Systems in the Era of Large Language Models ({{LLMs}})},
  author = {Zhao, Zihuai and Fan, Wenqi and Li, Jiatong and Liu, Yunqing and Mei, Xiaowei and Wang, Yiqi and Wen, Zhen and Wang, Fei and Zhao, Xiangyu and Tang, Jiliang and Li, Qing},
  year = {2024},
  month = nov,
  journal = {IEEE Transactions on Knowledge and Data Engineering},
  volume = {36},
  number = {11},
  pages = {6889--6907},
  issn = {1041-4347, 1558-2191, 2326-3865},
  doi = {10.1109/TKDE.2024.3392335},
  urldate = {2025-10-02},
  copyright = {https://ieeexplore.ieee.org/Xplorehelp/downloads/license-information/IEEE.html},
  langid = {english}
}

@misc{llmrigidity2025,
  author       = {{Liliz Lab}},
  title        = {LLM Rigidity Bandits},
  year         = {2025},
  howpublished = {\url{https://github.com/Liliz-lab/llm-rigidity-bandits}},
  note         = {Accessed: 2025-10-02}
}

@article{nord-bronzykAssessingRiskImplementing2025,
  title = {Assessing Risk in Implementing New Artificial Intelligence Triage Tools---How Much Risk Is Reasonable in an Already Risky World?},
  author = {{Nord-Bronzyk}, Alexa and Savulescu, Julian and Ballantyne, Angela and {Braunack-Mayer}, Annette and Krishnaswamy, Pavitra and Lysaght, Tamra and Ong, Marcus E. H. and Liu, Nan and Menikoff, Jerry and Mertens, Mayli and Dunn, Michael},
  year = {2025},
  month = jan,
  journal = {Asian Bioethics Review},
  volume = {17},
  number = {1},
  pages = {187--205},
  issn = {1793-8759, 1793-9453},
  doi = {10.1007/s41649-024-00348-8},
  urldate = {2025-10-05},
  langid = {english}
}

@article{challenArtificialIntelligenceBias2019,
  title = {Artificial Intelligence, Bias and Clinical Safety},
  author = {Challen, Robert and Denny, Joshua and Pitt, Martin and Gompels, Luke and Edwards, Tom and {Tsaneva-Atanasova}, Krasimira},
  year = {2019},
  month = mar,
  journal = {BMJ Quality \& Safety},
  volume = {28},
  number = {3},
  pages = {231--237},
  issn = {2044-5415, 2044-5423},
  doi = {10.1136/bmjqs-2018-008370},
  urldate = {2025-10-05},
  langid = {english},
}




\end{document}